\documentclass[12pt]{article} % use larger type; default would be 10pt
\usepackage{amsfonts,amsmath,amssymb,amsthm,dsfont,epsfig,graphicx,booktabs,latexsym} %
\usepackage[authoryear,round]{natbib}
\usepackage{enumerate}
\usepackage{multirow}
\usepackage{color}
\usepackage{amssymb,mathrsfs}
\usepackage{amsbsy,amsthm}
\usepackage{float,epsfig,subfigure}
\usepackage{multirow,setspace}
\usepackage{authblk,epstopdf}
\usepackage{array} % for better arrays (eg matrices) in maths
\usepackage{paralist}
\usepackage{verbatim} % adds environment for commenting out blocks of text & for better verbatim
\usepackage{graphicx}
\usepackage{mathtools}
\usepackage{blkarray}
\usepackage{stackengine}
\usepackage{xcolor}
\usepackage{caption}
\usepackage{arydshln}

\usepackage[margin=1in]{geometry}
\newsavebox\CBox

\begin{document}
\title{Multimodal Deep Generative Model for Semi-Supervised Learning under Class Imbalance}
\date{}
%\author{}
\author[a]{Heegeon Yoon}
\author[a,+]{Heeyoung Kim}
\affil[a]{\small{Department of Industrial and Systems Engineering, Korea Advanced Institute of Science and Technology (KAIST), Daejeon, Republic of Korea}}
\affil[+]{Corresponding author, Email: heeyoungkim@kaist.ac.kr}
\maketitle
\begin{abstract}
When modeling class-imbalanced data, it is crucial to address the imbalance, as models trained on such data tend to be biased towards the majority classes. This problem is amplified under partial supervision, where pseudo-labels for unlabeled data are predicted based on imbalanced labeled data, propagating the bias. While recent semi-supervised models address class imbalance, they typically assume single-modal input data. However, with the growing availability of multimodal data, it is essential to leverage complementary modalities. In this article, we propose a multimodal deep generative model for semi-supervised learning under class imbalance. Our approach uses separate encoders for each modality, sharing latent variables across modalities, and simplifies joint posterior computation with a product-of-experts method. To further address class imbalance, we replace typical Gaussian distributions with Student’s $t$-distributions for the prior, encoder, and decoder, better capturing the heavy-tailed latent distributions in imbalanced data. 
We derive a new objective function for training the proposed model on both labeled and unlabeled data using $\gamma$-power divergence.
Empirical results on benchmark and real-world datasets demonstrate that our model outperforms baseline methods in generalization, achieving superior classification performance for partially labeled multimodal data with imbalanced class distributions.
\end{abstract}

\section{Introduction}
Deep neural networks (DNNs) have demonstrated remarkable effectiveness across a wide range of applications owing to their powerful computational capacity \citep{kim2021locally,kim2023contextual,yoon2024uncertainty}. As the volume of accessible data increases and input patterns become more intricate, DNNs have evolved to incorporate more complex structures and higher computational capabilities \citep{yoon2025uncertainty,koo2024deep}. However, despite their impressive performance, DNNs are prone to overfitting when trained on datasets with class imbalance \citep{lee2021abc,shu2019meta}. Specifically, DNNs tend to favor majority classes in such scenarios, leading to poor  generalization on minority classes \citep{lee2025learnable, park2024rebalancing, zhang2023deep}. This issue is exacerbated under partial supervision, where the pseudo-labels for unlabeled data—predicted based on a limited, class-imbalanced labeled dataset—further reinforce the imbalance during training \citep{lee2024cdmad, kim2020distribution}. Real-world datasets often complicate this further by being only partially labeled, due to the impracticality of fully annotating large datasets, and by exhibiting imbalanced class distributions, where minority classes frequently contain crucial information \citep{yoon2023label,lee2023semi,park2022prediction,ko2019fault}. Therefore, it is important to develop methods for robustly training DNNs on class-imbalanced data in semi-supervised learning settings, leveraging the abundance of unlabeled data while ensuring the preservation of crucial information from minority classes.

In addition, real-world datasets are often multimodal, meaning they combine information from diverse sources in multiple formats \citep{chung2019crime,choy2016looking}. To effectively utilize such data, various multimodal learning methods have been developed, leveraging the complementary information inherent in different modalities to enhance performance \citep{cho2023prediction, han2022multimodal,ko2022deep}. Discriminative multimodal learning approaches focus on bridging modalities by combining them  into a unified representation through various fusion techniques, ranging from simple concatenation of input features \citep{hong2020more, huang2021makes} to generating joint representations using neural network architectures \citep{li2021multi, wang2021mogonet}. In contrast, generative approaches \citep{wu2018multimodal, wu2019multimodal, shi2019variational} aim to model the joint distribution across modalities by exploring their latent representations, thereby capturing the semantic correlations between them.

Although numerous studies have addressed either class imbalance or multimodality in data, few have tackled the challenging scenario where datasets are both class-imbalanced and multimodal, and none have assumed partial supervision. However, addressing the task of class-imbalanced and multimodal learning under semi-supervised settings remains an important and unsolved problem across various domains. For example, in medical diagnosis, multiple types of data, such as X-ray images and patient meta-information (e.g., gender, age, and medical history), are often used to detect diseases. Simultaneously, the number of healthy patients typically far exceeds the number of infected ones, leading to class imbalance. Additionally, some patients may have their medical data collected but remain undiagnosed, meaning their infection status is unknown. This scenario highlights the need for solutions that can handle class imbalance and multimodality under partial supervision.

One might consider using zero-shot multimodal foundation models \citep{radford2021learning, li2022blip, li2023blip} for such tasks, given their impressive performance across a wide range of applications without task-specific training. However, these models are typically trained on large-scale, general-purpose datasets and often struggle to generalize to domain-specific tasks involving partial supervision and class imbalance. In addition, they tend to require substantial computational resources and offer limited interpretability. These limitations highlight the need for a more structured, data-efficient approach that explicitly models interactions among modalities, class labels, and latent factors under partial supervision.

In this article, we propose a unified framework for semi-supervised learning with class-imbalanced multimodal data. Specifically, we introduce a novel method called the semi-supervised multimodal variational autoencoder for class-imbalanced learning (SSMVAE-CI), which models the joint distribution of multiple modalities under imbalanced class distributions.  Our approach builds on the semi-supervised variational autoencoder (VAE) framework \citep{kingma2014semi}, which effectively utilizes unlabeled data by introducing latent variables for both observations and their corresponding labels. To accommodate multimodal data, we design modality-specific encoders that share a common latent space for each  latent variable. This shared latent space unifies heterogeneous information across modalities, enabling  better capture of inter-modal correlations. Additionally, to address class imbalance, we replace the commonly used Gaussian distribution with a Student's $t$-distribution for each encoder. The heavy-tailed nature of the $t$-distribution enhances the model's ability to effectively represent imbalanced class distributions.

Our design of modality-specific encoders addresses the challenge of exponential growth in trainable parameters that arises with separate inference networks for each combination of modalities. We mathematically formulate the relationship between the joint posterior and the individual posteriors of the modalities using a product-of-experts (PoE) framework, inspired by the multimodal variational autoencoder (MVAE) \citep{wu2018multimodal}. Furthermore, our use of the Student’s $t$-distribution in each encoder addresses class imbalance by enhancing the flexibility of the latent space. Compared to the Gaussian, the $t$-distribution exhibits heavier tails, maintaining non-negligible density in sparse regions of the latent space where minority-class samples typically reside due to their scarcity. This reduces the risk of over-regularizing these samples during variational inference, allowing the encoder to preserve semantically meaningful but underrepresented patterns. Prior work such as $t^3$VAE \citep{kim2023t} has explored the use of Student's $t$-distributions in VAEs to enhance robustness, providing inspiration for our heavy-tailed posterior design. However, $t^3$VAE is restricted to unimodal and unsupervised settings. A similar motivation underlies the tilted VAE proposed by \cite{floto2023tilted}, which modifies the Gaussian prior to allocate more probability mass to the tails, thereby better supporting latent encodings that fall in low-density regions. While their approach is not specifically designed to address class imbalance, it shares the underlying goal of improving representation quality for samples that deviate from high-density regions. In our case, the $t$-distribution achieves a similar effect through its inherent heavy-tailed nature, enabling more faithful posterior representations for minority-class samples in class-imbalanced scenarios.

Unlike MVAE, which assumes only continuous latent variables in an unsupervised setting, and $t^3$VAE, which is limited to unimodal and unsupervised scenarios, our approach introduces discrete latent class variables and a multimodal extension of the Student's $t$-distribution. This enables joint modeling of continuous and discrete latent variables across multiple modalities under partial supervision. By integrating these elements via a PoE inference mechanism, our framework resolves incompatibilities in supervision levels, latent variable types, and modality handling, rather than simply combining existing methods. This unified probabilistic formulation ensures theoretical consistency for semi-supervised multimodal learning while maintaining robustness to class imbalance, allowing the model to capture complex relationships that neither MVAE nor $t^3$VAE can address individually. Although our formulation does not offer formal convergence or generalization guarantees, deriving such theoretical results remains challenging given the heterogeneous probabilistic components. Instead, we provide extensive empirical validation across diverse datasets and evaluation settings, demonstrating consistently strong performance and suggesting that the model generalizes well in practice.

To facilitate end-to-end optimization of the encoder and decoder parameters, we derive a new objective function for both labeled and unlabeled data points. This is achieved by solving for a closed-form expression of the $\gamma$-power divergence between two joint distributions---one representing the model distribution manifold and the other representing the data distribution manifold--—over the observed variable, continuous latent variable, and latent class variable. This divergence-based formulation ensures both robustness and expressiveness in the learned representations, particularly in partially labeled, class-imbalanced, and multimodal settings.

Our contributions are summarized as follows:\\
(i) We propose the first generative model that jointly tackles semi-supervision, multimodality, and class imbalance within a unified framework. These three challenges are typically addressed separately due to differences in modeling requirements (e.g., label handling, modality fusion, distributional assumptions), and our model is carefully designed to reconcile these differences in a principled manner. \\
(ii) We develop a PoE inference mechanism that supports both continuous and discrete latent variables across modalities, enabling flexible and scalable multimodal modeling. \\
(iii) We generalize the use of the Student’s $t$-distribution to the multimodal setting, enhancing robustness to class imbalance and improving tail-class modeling through heavy-tailed posterior approximations. \\
(iv) We derive a new training objective based on $\gamma$-power divergence, which ensures stable and expressive learning from both labeled and unlabeled data under imbalanced class distributions. \\
(v) We conduct extensive experiments on multiple datasets, demonstrating that our approach consistently outperforms baseline methods in semi-supervised, multimodal, and class-imbalanced settings.

The remainder of this paper is organized as follows. Section \ref{sec:review} reviews related work. Section \ref{sec:background} provides a brief overview of deep generative models and their extensions to semi-supervised learning, multimodal learning, and class-imbalanced learning. In Section \ref{sec:method}, we detail the proposed model and its inference process. Section \ref{sec:experiment} presents experimental results on benchmark and real-world datasets. Finally, Section \ref{sec:conclusion} concludes the paper.

\section{Related Work}\label{sec:review}
\subsection{Semi-Supervised Learning}
Recent methods in semi-supervised learning (SSL) are mostly based on consistency regularization \citep{laine2016temporal, tarvainen2017mean, miyato2018virtual}, which encourages models to produce consistent class distributions for an unlabeled example and its augmented version. Another prominent approach is pseudo-labeling \citep{lee2013pseudo}, where pseudo-labels are generated for unlabeled samples and used for model prediction. Holistic methods \citep{berthelot2019mixmatch, berthelot2019remixmatch, sohn2020fixmatch} further improve these techniques by combining multiple strategies. In contrast to these  discriminative approaches, generative methods focus on learning the data distribution to produce predictions. Typical examples include models based on VAEs \citep{kingma2014semi, louizos2015variational, ehsan2017infinite, davidson2018hyperspherical, feng2021shot} and generative adversarial networks (GANs) \citep{odena2016semi, liu2020catgan}. This paper mainly focuses on VAE-based generative models, specifically the semi-supervised VAE, which supports scalable inference and data generation. By encoding input data into a lower dimensional latent representation, this approach enables efficient modeling of large-scale, high-dimensional datasets, where dimensionality reduction is essential \citep{soh2018application,lee2013dependence}.

\subsection{Class-Imbalanced Learning}
Two common approaches for class-imbalanced learning are re-sampling \citep{chawla2002smote,liu2008exploratory, he2009learning, byrd2019effect} and re-weighting \citep{wang2017learning, ross2017focal, khan2017cost, ren2018learning, hu2019learning, cui2019class, cao2019learning}. Re-sampling methods balance class distributions by under-sampling the majority class or over-sampling the minority class, while re-weighting methods adjusts loss functions by assigning greater weights to minority samples. Generative oversampling \citep{wan2017variational, zhang2018over, mullick2019generative, wang2020deep, ai2023generative} extends re-sampling methods by generating synthetic samples from learned data distributions, improving scalability. Other generative approaches \citep{takahashi2018student, abiri2020variational, mathieu2019disentangling, kim2023t} incorporate Student's $t$-distributions, leveraging its heavy-tailed properties.  This paper focuses on employing the $t$-distribution to define the generative model in a semi-supervised VAE designed for SSL to better capture the heavy-tailed nature of class-imbalanced data.

\subsection{Multimodal Learning}
Typical multimodal learning models utilize multimodal fusion techniques to integrate information from different modalities. These fusion techniques are categorized into early fusion \citep{poria2015deep, poria2016convolutional, wang2017select}, intermediate fusion \citep{zadeh2017tensor, liu2018efficient, hou2019deep, han2022multimodal}, and late fusion \citep{nojavanasghari2016deep, wortwein2017really, tian2020uno}, based on the stage at which the fusion occurs. Dynamic fusion \citep{xue2023dynamic, zhang2023provable, cao2024predictive} introduces instance-dependent weights for output aggregation, enabling modality-level dynamics and enhancing the interpretability of late fusion methods. Meanwhile, generative approaches \citep{wu2018multimodal, wu2019multimodal, shi2019variational, lee2021private}  employ the VAE framework to model joint distributions across modalities, offering fast and scalable inference. The proposed model distinguishes itself from these generative approaches by incorporating a latent class variable to account for the semi-supervised nature of the input data and employing heavy-tailed distributions in the latent space to better represent imbalanced datasets.

\section{Background}\label{sec:background}
\subsection{Deep Generative Models for Semi-Supervised Learning and Multimodal Learning}\label{sec:mvae}
The semi-supervised VAE \citep{kingma2014semi} and the MVAE \citep{wu2018multimodal} extend the VAE \citep{kingma2013auto} for semi-supervised and multimodal learning, respectively. The semi-supervised VAE introduces a discrete class variable $y$, latent only for unlabeled data, while the MVAE models $M$ modalities $\mathbf{x}^1,...,\mathbf{x}^M$, assumed conditionally independent given a shared latent variable $\mathbf{z}$.  Both models share the prior $p(\mathbf{z})=\mathcal{N}(\mathbf{z}|\mathbf{0},\mathbf{I})$, but differ in their likelihood functions: the semi-supervised VAE conditions its likelihood on $y$ with $p(y)=\text{Cat}(y|\mathbf{\pi})$, where $\text{Cat}(y|\mathbf{\pi})$ is a multinomial distribution, while the MVAE's likelihood factors across modalities as
\begin{equation}\label{eq.1}
  p_\theta (\mathbf{x}^1,...,\mathbf{x}^M|\mathbf{z})
  =\prod_{m=1}^M p_\theta (\mathbf{x}^m|\mathbf{z}).
\end{equation}

Inference models $q_\phi(\cdot)$ approximate the posterior for both models. The semi-supervised VAE assumes a factorized form $q_\phi (y,\mathbf{z}|\mathbf{x})=q_\phi (\mathbf{z}|\mathbf{x},y)q_\phi (y|\mathbf{x})$, 
while the MVAE employs a PoE approach that uses the relation $p(\mathbf{z}|\mathbf{x}^1,...,\mathbf{x}^M) \propto \frac{\prod_{m=1}^M p(\mathbf{z}|\mathbf{x}^m)}{\prod_{m=1}^{M-1} p(\mathbf{z})}$ to approximate the joint posterior $q_\phi(\mathbf{z}|\mathbf{x}^1,...,\mathbf{x}^M)$, reducing the need for $2^M$ inference networks: 
\begin{equation}\label{eq.2}
    q_\phi(\mathbf{z}|\mathbf{x}^1,...,\mathbf{x}^M) \approx \frac{\prod_{m=1}^M q_\phi(\mathbf{z}|\mathbf{x}^m)}{\prod_{m=1}^{M-1} p(\mathbf{z})}\equiv \frac{\prod_{m=1}^M [\tilde{q}_\phi (\mathbf{z}|\mathbf{x}^m)p(\mathbf{z})]}{\prod_{m=1}^{M-1} p(\mathbf{z})}=p(\mathbf{z})\prod_{m=1}^M \tilde{q}_\phi (\mathbf{z}|\mathbf{x}^m),
\end{equation}
where $q_\phi (\mathbf{z}|\mathbf{x}^m)\equiv \tilde{q}_\phi (\mathbf{z}|\mathbf{x}^m)p(\mathbf{z})$ with $\tilde{q}_\phi (\mathbf{z}|\mathbf{x}^m)$ being the underlying inference network.

Using the inference models, the evidence lower bounds (ELBOs) on the marginal likelihood are derived for both models. For the semi-supervised VAE, the ELBOs are computed separately for labeled and unlabeled data: for labeled data, the ELBO extends that of the original VAE by including the observed class variable $y$, while for unlabeled data, $y$ is treated as latent. 
In the MVAE, the ELBO generalizes the original VAE by replacing the single-modality input $\mathbf{x}$ with  $M$ modalities $\mathbf{x}^1,...,\mathbf{x}^M$. For further details, see \cite{kingma2014semi} and \cite{wu2018multimodal}.

\subsection{VAE Framework for Heavy-tailed Data}\label{sec:t3vae}
To model the heavy-tailed nature of real-world datasets, the $t^3$VAE \citep{kim2023t} extends the VAE framework by incorporating Student's $t$-distributions. The generative process assumes a latent variable $\mathbf{z}\in\mathbb{R}^{n_\mathbf{z}}$ for observed data $\mathbf{x}\in\mathbb{R}^{n_\mathbf{x}}$ with: 
\begin{equation}\label{eq.3}
    p_{\nu}(\mathbf{z})=t_{n_\mathbf{z}}(\mathbf{z}|\mathbf{0},\mathbf{I},\nu), \qquad p_{\theta,\nu}(\mathbf{x}|\mathbf{z})=t_{n_\mathbf{x}}\left(\mathbf{x}\bigg|\mathbf{\mu}_{\theta}(\mathbf{z}),\frac{1+\nu^{-1}\|\mathbf{z}\|^2}{1+\nu^{-1}n_\mathbf{z}}\sigma^2 \mathbf{I},\nu+n_\mathbf{z}\right),
\end{equation}
where $\nu>2$ denotes the degrees of freedom.
The corresponding inference model is $q_{\phi,\nu}(\mathbf{z}|\mathbf{x})=t_{n_\mathbf{z}}(\mathbf{z}|\mathbf{\mu}_\phi(\mathbf{x}),(1+\nu^{-1}n_\mathbf{x})^{-1}\mathbf{\Sigma}_\phi(\mathbf{x}),\nu+n_\mathbf{x})$, which is derived from the true posterior distribution for the case where the mean $\mathbf{\mu}_{\theta}(\mathbf{z})$ of the likelihood $p_{\theta,\nu}(\mathbf{x}|\mathbf{z})$ is linear.

To optimize the model, the $t^3$VAE employs the $\gamma$-loss, derived from the $\gamma$-power divergence, suitable for power-family distributions. 
The closed-form $\gamma$-loss is given by 
\begin{multline}\label{eq.4}
\mathcal{J}_\gamma(\theta,\phi)=\frac{1}{2}\mathbb{E}_{\mathbf{x}\sim p_{\text{data}}}\left[\frac{1}{\sigma^2}\mathbb{E}_{\mathbf{z}\sim q_{\phi,\nu}(\cdot|\mathbf{x})}\|\mathbf{x}-\mathbf{\mu}_\theta(\mathbf{z})\|^2\right. \\
+\left.\|\mathbf{\mu}_\phi(\mathbf{x})\|^2+\frac{\nu}{\nu+n_\mathbf{x}-2}\text{tr}\mathbf{\Sigma}_\phi(\mathbf{x})-\frac{\nu C_1}{C_2}|\mathbf{\Sigma}_\phi(\mathbf{x})|^{-\frac{\gamma}{2(1+\gamma)}}\right],
\end{multline}
where $C_1$ and $C_2$ are constants dependent on $\nu$, $n_\mathbf{z}$, and $n_\mathbf{x}$. 
For further details on the $t^3$VAE, see \cite{kim2023t}.

\section{Methodology}\label{sec:method}

In this section, we propose a new deep generative model for semi-supervised learning with class-imbalanced multimodal data. In Section \ref{sec:problem}, we present the problem scenario, followed by a description of the proposed model in Section \ref{sec:generative} and its inference process in Section \ref{sec:inference}.

\subsection{Problem Scenario}\label{sec:problem}

Our training dataset consists of both labeled and unlabeled data from multiple modalities. The labeled dataset is given as $\mathcal{D}_{L}=\{(\{\mathbf{x}_1^1,...,\mathbf{x}_1^M\},y_1),...,(\{\mathbf{x}_L^1,...,\mathbf{x}_L^M\},y_L)\}$, where $\mathbf{x}_i^m$ represents the $i$th instance of the $m$th modality, and $y_i\in\{1,...,K\}$ denotes the corresponding label. The unlabeled dataset is denoted as $\mathcal{D}_{U}=\{\{\mathbf{x}_{L+1}^1,...,\mathbf{x}_{L+1}^M\},... ,\{\mathbf{x}_{L+U}^1,...,\mathbf{x}_{L+U}^M\}\}$. In total, the dataset contains $L+U$ samples, where class $k$ has $L_k$ labeled samples, such that $L=\sum_{k=1}^K L_k$. We assume an imbalanced class distribution, where class sizes differ  significantly. If classes are sorted by size in descending order ($L_i\geq L_j$ for $i<j$), the largest class ($L_1$) dominates the smallest class ($L_K$), with the imbalance ratio defined as $r=L_1/L_K$. The goal of our model is to predict the class label $y$ of an unseen instance $\{\mathbf{x}^1,...,\mathbf{x}^M\}$ using both $\mathcal{D}_L$ and $\mathcal{D}_U$.

\subsection{Proposed Model}\label{sec:generative}

We propose SSMVAE-CI, a novel multimodal deep generative model for semi-supervised learning with imbalanced class distributions. 

We assume that each data point comprises $M$ modalities $\mathbf{x}^1,\ldots,\mathbf{x}^M$, which share a common latent variable $\mathbf{z}$ and a class variable $y$. For unlabeled data, $y$ is treated as a latent variable. Additionally, we assume that the $M$ modalities are conditionally independent given $\mathbf{z}$. Under this assumption, we extend the generative process of the semi-supervised VAE to incorporate $M$ modalities. Specifically, the conditional independence assumption allows the joint likelihood to be expressed as a product of individual likelihoods, such that Equation \eqref{eq.1} becomes:
\begin{equation}\label{eq.5}
    p_\theta (\mathbf{x}^1,...,\mathbf{x}^M|y,\mathbf{z})=\prod_{m=1}^M p_\theta (\mathbf{x}^m|y,\mathbf{z}),
\end{equation}
where each individual likelihood $p_\theta (\mathbf{x}^m|y,\mathbf{z})=f(\mathbf{x}^m;y,\mathbf{z},\mathbf{\theta})$ is modeled by a neural network with parameters $\mathbf{\theta}$.

The generative process of SSMVAE-CI consists of a prior for the class variable $y$, a prior for the latent variable $\mathbf{z}$, and the joint likelihood conditioned on these two variables. 
To address class imbalance in the training data, the prior for $\mathbf{z}$ is assumed to follow a $t$-distribution, instead of a typical Gaussian distribution, with mean $\mathbf{0}$, scale matrix $\mathbf{I}$, and $\nu>2$ degrees of freedom, i.e., $p_\nu(\mathbf{z})=t_{n_z}(\mathbf{z}|\mathbf{0},\mathbf{I},\nu)$, where $n_\mathbf{z}$ is the dimensionality of $\mathbf{z}$. The joint likelihood is expressed as a product of individual likelihoods, as in Equation \eqref{eq.5}. Each individual likelihood,  $p_{\theta,\nu}(\mathbf{x}^m|y,\mathbf{z})=f(\mathbf{x}^m;y,\mathbf{z},\theta,\nu)$, for modalities $m=1,...,M$, is also modeled as a $t$-distribution and is parameterized by a neural network.

\subsection{Inference}\label{sec:inference}

To address the intractability of the latent variables $y$ and $\mathbf{z}$ in SSMVAE-CI, we approximate the true joint posterior $p(y,\mathbf{z}|\mathbf{x}^1,...,\mathbf{x}^M)$ by introducing an approximate posterior distribution $q_\phi(\cdot)$, parameterized by $\phi$ as an inference model, inspired by the semi-supervised VAE. We further assume a factorized form for the approximate posterior:  $q_\phi(y,\mathbf{z}|\mathbf{x}^1,...,\mathbf{x}^M)=q_\phi(y|\mathbf{x}^1,...,\mathbf{x}^M)q_\phi(\mathbf{z}|\mathbf{x}^1,...,\mathbf{x}^M,y)$, where $q_\phi(y|\mathbf{x}^1,...,\mathbf{x}^M)$ is modeled using a multinomial distribution, and $q_\phi(\mathbf{z}|\mathbf{x}^1,...,\mathbf{x}^M,y)$ is modeled using a $t$-distribution. To reduce the number of inference networks required for modeling $q_\phi (\mathbf{z}|\mathbf{x}^1,...,\mathbf{x}^M,y)$, we employ a PoE framework with a prior expert \citep{wu2018multimodal}. This allows us to express the approximate joint posterior as the product of the approximate posteriors for individual modalities. Specifically, the approximate joint posterior can be factorized as follows:
\begin{eqnarray}
    q_\phi (y,\mathbf{z}|\mathbf{x}^1,...,\mathbf{x}^M)&=&q_\phi (y|\mathbf{x}^1,...,\mathbf{x}^M) q_\phi (\mathbf{z}|\mathbf{x}^1,...,\mathbf{x}^M,y) \nonumber\\
    &=&q_\phi (y|\mathbf{x}^1,...,\mathbf{x}^M)p(\mathbf{z})\prod_{m=1}^M q_\phi (\mathbf{z}|\mathbf{x}^m,y),\label{eq.6}
\end{eqnarray}
where the individual posterior $q_\phi(\mathbf{z}|\mathbf{x}^m,y)$ is modeled as a $t$-distribution, i.e., $q_{\phi,\nu}(\mathbf{z}|\mathbf{x}^m,y)=t_{n_\mathbf{z}}(\mathbf{z}|\mu_\phi(\mathbf{x}^m,y),\tilde{\Sigma}_\phi(\mathbf{x}^m,y),\nu+n_{\mathbf{x}^m}+K)$, where $\mu_\phi(\mathbf{x}^m,y)$ and $\tilde{\Sigma}_\phi(\mathbf{x}^m,y)=(1+\nu^{-1}(n_{\mathbf{x}^m}+K))^{-1}\Sigma_\phi(\mathbf{x}^m,y)$ denote the mean and the scale matrix of the $t$-distribution, respectively, with $n_{\mathbf{x}^m}$ being the dimensionality of $\mathbf{x}^m$.

To update the parameters $\theta$ and $\phi$ of SSMVAE-CI, we construct an objective function using the $\gamma$-power divergence \citep{kim2023t}. In developing this objective function, we separately consider two cases: one for labeled data points and another for unlabeled data points, following the approach of the semi-supervised VAE. 
In the first case, where the label of a data point is observed, the objective function for a single modality $m\in \{1,...,M\}$ is an extension of the original $\gamma$-loss function in Equation \eqref{eq.4}, incorporating the class variable $y$:
\begin{multline}\label{eq.7}
\mathcal{L}_\gamma^m(\theta,\phi)=\frac{1}{2}\mathbb{E}_{(\mathbf{x}^m,y)\sim\mathcal{D}_L}\left[\frac{1}{\sigma^2}\mathbb{E}_{\mathbf{z}\sim q_{\phi,\nu}(\cdot|\mathbf{x}^m,y)}\|\mathbf{x}^m-\mathbf{\mu}_\theta(y,\mathbf{z})\|^2\right. \\
 +\left.\|\mathbf{\mu}_\phi(\mathbf{x}^m,y)\|^2+\frac{\nu}{\nu+n_{\mathbf{x}^m}+K-2}\text{tr}\mathbf{\Sigma}_\phi(\mathbf{x}^m,y)-\frac{\nu C_1'}{C_2'}|\mathbf{\Sigma}_\phi(\mathbf{x}^m,y)|^{-\frac{\gamma}{2(1+\gamma)}}\right],
\end{multline}
where $\gamma$ is related to $\nu$ as $\gamma=-\frac{2}{\nu+n_{\mathbf{x}^m}+K+n_\mathbf{z}}$, and $C_1'$ and $C_2'$ are constants dependent on $\nu$, $n_\mathbf{z}$, $n_{\mathbf{x}^m}$, and $K$. Since there are $M$ such objective functions, one for each modality, the overall objective function across all $M$ modalities is given by $\mathcal{L}_\gamma(\theta,\phi)=\sum_{m=1}^M \mathcal{L}_\gamma^m(\theta,\phi)$. This summation is justified by the fact that both the joint likelihood in Equation \eqref{eq.5} and the joint posterior in Equation \eqref{eq.6} can be decomposed into $M$ distinct components.

In the second case, where the label is missing, it is treated as a latent variable, and the objective function for a data point with an unobserved label $y$ is given by:
\begin{equation}
\begin{split}
\mathcal{U}_\gamma(\mathbf{x}^1,...,\mathbf{x}^M;\theta,\phi)=&\: \mathbb{E}_{\substack{\mathbf{z}\sim q_{\phi,\nu}(\cdot|\mathbf{x}^1,...,\mathbf{x}^M,y) \\ y\sim q_\phi(\cdot|\mathbf{x}^1,...,\mathbf{x}^M)}}\left[\frac{1}{2\sigma^2} \|(\mathbf{x}^1,...,\mathbf{x}^M)-\mathbf{\mu}_\theta(y,\mathbf{z})\|^2\right] \\
    &  +\mathcal{D}_\gamma(q_{\phi,\nu}(y,\mathbf{z}|\mathbf{x}^1,...,\mathbf{x}^M)||p(y,\mathbf{z})) \\
    =&\sum_y q_\phi (y|\mathbf{x}^1,...,\mathbf{x}^M)\mathcal{L}_\gamma(\mathbf{x}^1,...,\mathbf{x}^M,y;\theta,\phi) + C_1' \mathcal{H}(q_\phi (y|\mathbf{x}^1,...,\mathbf{x}^M)),\label{eq.8}
\end{split}
\end{equation}
where $C_1'$ is the same constant as in Equation \eqref{eq.7}, and $\mathcal{H}$ denotes the entropy. The derivation of this objective function is motivated by the structure of Equation \eqref{eq.7}, which consists of the reconstruction error (described by the first term in the bracket) and the regularizer for the encoder $q_{\phi,\nu}(\mathbf{z}|\mathbf{x}^m,y)$, represented by the remaining terms. A full derivation of the objective function in Equation \eqref{eq.8} is provided in the supplementary materials.

This divergence-based formulation can be intuitively understood as redefining the geometric relationship between the model and data manifolds. Unlike the KL divergence, which enforces strict overlap, the $\gamma$-power divergence introduces a controllable robustness parameter that allows smoother alignment between the two, reducing sensitivity to outliers and class imbalance. For labeled data, this objective coincides with the $\gamma$-loss of $t^3$VAE, except that we additionally condition on the class variable $y$. 
For unlabeled data, however, our approach differs fundamentally from $t^3$VAE: by treating $y$ as a latent variable and marginalizing it out through the $\gamma$-divergence, we derive a tractable lower bound analogous to the ELBO, enabling consistent optimization in the absence of labels. This unified treatment extends the $\gamma$-divergence framework of $t^3$VAE---originally limited to unimodal and fully unsupervised settings---into a semi-supervised, multimodal framework that preserves its inherent robustness to class imbalance while enabling flexible learning across modalities and supervision levels.

Finally, a classification loss is added to encourage the label predictive distribution $q_\phi (y|\mathbf{x}^1,...,\mathbf{x}^M)$ to learn effectively from the labeled data. Incorporating this classification loss, the overall objective function becomes:
\begin{equation}
\mathcal{J}_{overall}=\mathcal{L}_\gamma(\theta,\phi) + \sum_{(\mathbf{x}^1,...,\mathbf{x}^M) \sim \mathcal{D}_U}\mathcal{U}_\gamma(\mathbf{x}^1,...,\mathbf{x}^M;\theta,\phi)
- \alpha \cdot \mathbb{E}_{\mathcal{D}_L}\left[\log q_\phi (y|\mathbf{x}^1,...,\mathbf{x}^M)\right],\label{eq.9}
\end{equation}
where $\alpha$ controls the contribution of the discriminative learning from the label predictive distribution $q_\phi(y|\mathbf{x}^1,...,\mathbf{x}^M)$ during the training process.

Following the semi-supervised VAE framework \citep{kingma2014semi}, the hyperparameter $\alpha$ in Equation \eqref{eq.9} controls the balance between the generative term, which promotes coherent latent representations, and the discriminative term, which enhances classification performance. Smaller $\alpha$ values favor generative learning, whereas larger values emphasize discriminative learning. In our experiments, we found that using a moderately larger value ($\alpha = 10 \times N$, where $N=L+U$) yielded slightly smoother and more stable convergence under class imbalance, while overall performance remained consistent across different $\alpha$ settings. This suggests that the model is not overly sensitive to $\alpha$, and that stronger discriminative weighting can help stabilize training in more complex, imbalanced scenarios. Other hyperparameters, such as the latent dimensionality $n_\mathbf{z}$ and the degrees of freedom $\nu$, were set to the values commonly used in prior VAE and $t^3$VAE studies, which also produced stable and reliable results in our experiments.

The reparameterization trick for backpropagation of gradients through stochastic nodes must be adapted to enable sampling $\mathbf{z}_{(i,1)},...,\mathbf{z}_{(i,L)}$ from the approximate posterior distribution $q_{\phi,\nu}(\mathbf{z}|\mathbf{x}^m,y)$ for each data pair $(\mathbf{x}_{(i)}^m,y_{(i)})$. Inspired by the $t$-reparameterization trick \citep{kim2023t}, which generates samples from a multivariate $t$-distribution $T\sim t_d(\mu,\Sigma,\nu)$ using an affine transformation of a multivariate centered Gaussian $Z\sim \mathcal{N}_d(0,\Sigma)$ and an independent chi-squared variable $V\sim \chi^2(\nu)$ via the relation $T\overset{d}{=}\mu+\frac{Z}{\sqrt{V/\nu}}$, we derive a modified $t$-reparameterization trick tailored for SSMVAE-CI. The trick is formulated as follows:
\begin{eqnarray} \mathbf{z}_{(i,l)}&=&\mu_\phi(\mathbf{x}_{(i)}^m,y_{(i)})+{\frac{1}{\sqrt{\delta_{(i,l)}/(\nu+n_{\mathbf{x}^m}+K)}}}{\frac{\sigma_\phi(\mathbf{x}_{(i)}^m,y_{(i)})}{\sqrt{1+\nu^{-1}(n_{\mathbf{x}^m}+K)}}} \odot \epsilon_{(i,l)} \nonumber \\
  &=&\mu_\phi(\mathbf{x}_{(i)}^m,y_{(i)})+\sqrt{\frac{\nu}{\delta_{(i,l)}}}\sigma_\phi(\mathbf{x}_{(i)}^m,y_{(i)}) \odot \epsilon_{(i,l)},\label{eq.10}
\end{eqnarray}
where $\epsilon_{(i,l)}\sim \mathcal{N}_{n_\mathbf{z}}(\mathbf{0},\mathbf{I})$ and $\delta_{(i,l)}\sim \chi^2(\nu+n_{\mathbf{x}^m}+K)$ are drawn independently.

Compared with standard Gaussian reparameterization, this $t$-based formulation not only enables low-variance gradient estimation but also preserves the heavy-tailed nature of the latent distribution. The resulting latent space allocates non-negligible probability mass to sparse or minority-class regions, preventing such samples from being over-regularized toward high-density areas. While $t^3$VAE employs the same idea only for continuous latent variables in a unimodal, unsupervised setting, our model extends the reparameterization to jointly accommodate both continuous and discrete latent variables across multiple modalities. This allows the model to maintain robustness and generalization under partial supervision, particularly in the presence of class imbalance.

\section{Experiments}\label{sec:experiment} 
In this section, we evaluate the effectiveness of our method through experiments on multiple multimodal datasets from various application domains. We begin by describing the experimental setup, including the datasets,  performance metrics, baseline methods, and implementation details. We then present and discuss the results.

\subsection{Experimental Setup}\label{sec:setup}

\textbf{Datasets.} In the first experiment, we used a combination of two benchmark image datasets:  the Modified National Institute of Standards and Technology (MNIST) dataset and the Street View House Numbers (SVHN) dataset. The MNIST dataset comprises 60,000 training and 10,000 test images of handwritten digits across 10 classes, while the SVHN dataset contains 73,257 training and 26,032 test images of house numbers, also across 10 classes. To create a multimodal dataset, we selected an equal number of images from each dataset for every class and paired one MNIST image with one SVHN image per label. 

In the second experiment, we employed the University Pierre and Marie Curie (UPMC) Food-101 dataset---a real-world multimodal dataset containing food images and corresponding textual recipe descriptions across 101 categories. Each class consists of 1,000 samples, with 750 training samples and 250 manually reviewed test samples. 

Lastly, we used the Carnegie Mellon University Multimodal Opinion Sentiment and Emotion Intensity (CMU-MOSEI) dataset---a real-world multimodal dataset consisting of over 23,500 sentence-utterance video clips from more than 1,000 speakers across 250 topics. Unlike the previous datasets, which involve two modalities, CMU-MOSEI includes three modalities: visual, audio, and textual.

For each dataset, we conducted extensive experiments with various imbalance ratios $r$ and different proportions of labeled data, denoted by $\beta=L/(L+U)$. Specifically, for the MNIST-SVHN and CMU-MOSEI datasets, we set $\beta$ to $10\%$ and $30\%$. For the MNIST-SVHN and UPMC Food-101 datasets, we varied the class imbalance ratio $r$ from 10 to 100 to evaluate the robustness of our method under different levels of class imbalance. To create imbalanced datasets, we followed the procedure described in \cite{cao2019learning} for generating imbalanced versions of the Canadian Institute for Advanced Research (CIFAR)-10 and CIFAR-100 datasets, where class sample sizes follow an exponential decay pattern. Across all datasets, we assumed that the class imbalance ratio of labeled data matched that of the unlabeled data and adjusted the number of unlabeled samples per class accordingly.

\textbf{Evaluation metrics.} To evaluate the performance of our model, we used two performance measures: overall classification accuracy and minority-class accuracy. Overall accuracy measures the proportion of correct predictions across the entire dataset, while minority-class accuracy measures the ratio of correct predictions only across the minority classes.
Specifically, minority-class accuracy is computed as the average per-class accuracy over the half of the classes with the fewest training samples. This metric highlights the model's robustness under class imbalance. Overall accuracy is computed as the ratio of correct predictions (true positives and true negatives) to the total number of samples.

\textbf{Baselines.} Since no existing methods simultaneously address multimodality, partial supervision, and class imbalance, we selected a diverse set of baseline approaches, each targeting a subset of these challenges. To enable fair comparisons, we additionally constructed composite baselines by combining existing class-imbalanced semi-supervised learning methods with multimodal fusion strategies---specifically, early and late fusion---to handle all three aspects.

We first considered methods developed for semi-supervised multimodal learning, which address both partial supervision and multimodality but do not explicitly handle class imbalance. These include: (i) Comprehensive Multi-Modal Learning (CMML) \citep{yang2019comprehensive}, which addresses modal insufficiency using an instance-level attention mechanism with diversity regularization and a robust consistency metric to preserve inter-modal diversity; and (ii) Semi-supervised Multimodal Learning Network (SMLN) \citep{hu2020semi}, which captures modality correlations by constructing a similarity matrix using both labeled and unlabeled data and optimizing a loss function based on spectral decomposition.

Next, we considered methods designed for class-imbalanced semi-supervised learning, which address class imbalance and partial supervision but assume unimodal inputs. These included: (i) Class Rebalancing Self-Training (CReST) \citep{wei2021crest}, which iteratively retrains a base SSL algorithm by adaptively incorporating pseudo-labeled samples; and (ii) Adaptive Thresholding (Adsh) \citep{guo2022class}, which dynamically adjusts the sampling of pseudo-labeled data based on class-dependent thresholds during training. To adapt these methods for multimodal scenarios, we applied early and late fusion strategies. In early fusion, raw features from all modalities are concatenated into a single input tensor, while in late fusion, predictions from individual modalities are aggregated. These composite versions therefore address all three challenges---multimodality, partial supervision, and class imbalance---within a modular design.

Finally, we included several methods that address only one of the three challenges, serving as reference points for evaluating the benefit of joint modeling. These include: (i) Semi-supervised VAE (SSVAE) \citep{kingma2014semi}, a generative model for semi-supervised learning that serves as the basis of our proposed framework; (ii) Label-Distribution-Aware Margin (LDAM) loss \citep{cao2019learning}, a re-weighting method for class-imbalanced learning that improves generalization through class-dependent margins; and (iii) Generative Adversarial Minority Oversampling (GAMO) \citep{mullick2019generative}, a Generative Adversarial Network (GAN)-based oversampling framework for class-imbalanced learning that involves a three-player game among a convex generator, classifier, and discriminator. As with other baselines not originally designed for multimodal data, we adapted these models using both early and late fusion strategies, following the procedures described above. Notably, LDAM and GAMO are originally designed for fully supervised settings. As our objective in including these models was to assess the added value of leveraging unlabeled data---rather than directly compare classification accuracy---we trained them using only the labeled portion of the dataset. Furthermore, although MVAE and $t^3$VAE are conceptually related to our framework, we did not include them as baselines because both are fundamentally unsupervised generative models. Adapting them for classification would require substantial modifications to their training objectives and inference procedures, making direct or fair comparison infeasible.

\textbf{Network structure and optimization.} For our model, we employed a 64-dimensional latent variable $\mathbf{z}$ and used ReLU as the activation function. The encoder and decoder networks for both modalities were implemented as multi-layer perceptrons (MLPs) with two hidden layers, each comprising 512 hidden units. During training, the class variable $y$ was concatenated with the instance $\mathbf{x}$ in the encoder networks to model the approximate posterior $q_\phi(\mathbf{z}|\mathbf{x},y)$,  and with the latent variable $\mathbf{z}$ in the decoder networks to model the likelihood $p_\theta(\mathbf{x}|y,\mathbf{z})$. The initial weight parameters were randomly sampled from $\mathcal{N}(\boldsymbol{0},0.001^2 \boldsymbol{I})$, while the initial bias parameters were set to 0. The objective function for the proposed model was optimized using the Adam with a batch size of 100, decay parameters for the first and second moments of the gradients $(\beta_1, \beta_2) = (0.9, 0.999)$ and a learning rate of $10^{-4}$.

\subsection{Results on MNIST-SVHN Dataset}\label{sec:benchmarks}
We repeated the experiment 10 times for each scenario; for each experiment, we randomly selected the required number of labeled and unlabeled images per class without replacement, ensuring the specified imbalance ratio was maintained. The classification accuracies, along with standard errors, are presented in Table \ref{table:accuracy_benchmarks}.

\begin{table}[ht]
\caption{Overall accuracies/minority-class accuracies for the MNIST-SVHN dataset.}
\label{table:accuracy_benchmarks}
\vskip 0.15in
\begin{center}
\begin{small}
\begin{sc}
    \resizebox{\textwidth}{!}{\begin{tabular}{ccccccc}
        \hline
        \multicolumn{2}{c}{Model} & $\beta (\%)$ & $r=10$ & $r=20$ & $r=50$ & $r=100$ \\ \hline
        \multicolumn{2}{c}{\multirow{2}{*}{CMML}} & 10 & 86.16{\scriptsize$\pm$0.06}/80.08{\scriptsize$\pm$0.13} & 80.26{\scriptsize$\pm$0.22}/62.94{\scriptsize$\pm$0.56} & 73.62{\scriptsize$\pm$0.49}/51.62{\scriptsize$\pm$0.55} & 64.99{\scriptsize$\pm$0.50}/41.80{\scriptsize$\pm$0.93} \\
        \multicolumn{2}{c}{} & 30 & 89.68{\scriptsize$\pm$0.44}/87.13{\scriptsize$\pm$0.71} & 84.56{\scriptsize$\pm$0.33}/72.20{\scriptsize$\pm$0.73} & 81.51{\scriptsize$\pm$0.20}/65.90{\scriptsize$\pm$0.42} & 69.37{\scriptsize$\pm$0.60}/51.17{\scriptsize$\pm$0.44} \\ \hline
        \multicolumn{2}{c}{\multirow{2}{*}{SMLN}} & 10 & 85.64{\scriptsize$\pm$0.37}/78.94{\scriptsize$\pm$0.55} & 80.51{\scriptsize$\pm$0.43}/60.84{\scriptsize$\pm$0.73} & 68.13{\scriptsize$\pm$0.59}/46.34{\scriptsize$\pm$0.88} & 61.61{\scriptsize$\pm$1.74}/39.34{\scriptsize$\pm$1.21} \\
        \multicolumn{2}{c}{} & 30 & 87.25{\scriptsize$\pm$0.93}/81.60{\scriptsize$\pm$1.08} & 81.02{\scriptsize$\pm$0.69}/68.76{\scriptsize$\pm$0.90} & 74.05{\scriptsize$\pm$1.51}/59.76{\scriptsize$\pm$1.97} & 63.78{\scriptsize$\pm$0.61}/43.26{\scriptsize$\pm$0.72} \\ \hline
        \multirow{4}{*}{SSVAE} & \multirow{2}{*}{+ Early fusion} & 10 & 89.36{\scriptsize$\pm$0.31}/84.30{\scriptsize$\pm$0.70} & 85.54{\scriptsize$\pm$0.35}/77.81{\scriptsize$\pm$0.71} & 78.82{\scriptsize$\pm$0.73}/64.59{\scriptsize$\pm$1.70} & 71.14{\scriptsize$\pm$1.24}/53.58{\scriptsize$\pm$1.69} \\
        & & 30 & 91.57{\scriptsize$\pm$0.39}/86.88{\scriptsize$\pm$0.62} & 88.45{\scriptsize$\pm$1.41}/81.84{\scriptsize$\pm$2.21} & 83.40{\scriptsize$\pm$0.90}/72.51{\scriptsize$\pm$1.57} & 74.55{\scriptsize$\pm$0.99}/56.49{\scriptsize$\pm$2.21} \\ \cline{2-7}
        & \multirow{2}{*}{+ Late fusion} & 10 & 75.63{\scriptsize$\pm$0.32}/67.52{\scriptsize$\pm$0.53} & 71.62{\scriptsize$\pm$1.19}/60.30{\scriptsize$\pm$0.97} & 66.04{\scriptsize$\pm$0.62}/50.29{\scriptsize$\pm$1.31} & 60.50{\scriptsize$\pm$0.59}/40.49{\scriptsize$\pm$1.26} \\
        & & 30 & 80.54{\scriptsize$\pm$0.71}/74.09{\scriptsize$\pm$1.48} & 77.30{\scriptsize$\pm$1.05}/67.99{\scriptsize$\pm$1.76} & 72.80{\scriptsize$\pm$1.08}/59.57{\scriptsize$\pm$1.18} & 69.02{\scriptsize$\pm$0.93}/53.13{\scriptsize$\pm$1.75} \\ \hline
        \multirow{4}{*}{GAMO} & \multirow{2}{*}{+ Early fusion} & 10 & 82.30{\scriptsize$\pm$0.99}/73.13{\scriptsize$\pm$1.70} & 78.55{\scriptsize$\pm$0.45}/62.55{\scriptsize$\pm$0.90} & 71.64{\scriptsize$\pm$0.54}/50.45{\scriptsize$\pm$1.50} & 64.63{\scriptsize$\pm$0.41}/36.13{\scriptsize$\pm$0.93} \\
        & & 30 & 86.30{\scriptsize$\pm$1.08}/79.41{\scriptsize$\pm$1.98} & 83.92{\scriptsize$\pm$1.11}/74.40{\scriptsize$\pm$2.58} & 79.27{\scriptsize$\pm$1.09}/64.40{\scriptsize$\pm$2.41} & 73.33{\scriptsize$\pm$1.17}/52.60{\scriptsize$\pm$1.84} \\ \cline{2-7}
        & \multirow{2}{*}{+ Late fusion} & 10 & 66.66{\scriptsize$\pm$0.82}/54.45{\scriptsize$\pm$1.46} & 60.81{\scriptsize$\pm$0.67}/43.59{\scriptsize$\pm$0.81} & 57.31{\scriptsize$\pm$0.32}/36.82{\scriptsize$\pm$0.53} & 52.45{\scriptsize$\pm$0.52}/28.58{\scriptsize$\pm$1.06} \\
        & & 30 & 70.18{\scriptsize$\pm$0.71}/60.88{\scriptsize$\pm$1.45} & 67.71{\scriptsize$\pm$0.74}/55.03{\scriptsize$\pm$1.27} & 64.61{\scriptsize$\pm$0.74}/48.31{\scriptsize$\pm$0.69} & 61.90{\scriptsize$\pm$1.18}/43.61{\scriptsize$\pm$1.85} \\ \hline
        \multirow{4}{*}{LDAM} & \multirow{2}{*}{+ Early fusion} & 10 & 83.45{\scriptsize$\pm$0.86}/74.70{\scriptsize$\pm$1.71} & 79.90{\scriptsize$\pm$0.93}/68.47{\scriptsize$\pm$1.80} & 73.51{\scriptsize$\pm$0.77}/54.72{\scriptsize$\pm$2.01} & 66.47{\scriptsize$\pm$0.93}/44.56{\scriptsize$\pm$3.06} \\
        & & 30 & 89.10{\scriptsize$\pm$1.02}/82.81{\scriptsize$\pm$2.34} & 87.02{\scriptsize$\pm$0.67}/80.39{\scriptsize$\pm$3.02} & 80.49{\scriptsize$\pm$0.37}/68.17{\scriptsize$\pm$2.20} & 74.40{\scriptsize$\pm$0.84}/56.68{\scriptsize$\pm$0.83} \\ \cline{2-7}
        & \multirow{2}{*}{+ Late fusion} & 10 & 73.28{\scriptsize$\pm$0.79}/66.18{\scriptsize$\pm$0.49} & 69.84{\scriptsize$\pm$0.62}/60.14{\scriptsize$\pm$1.55} & 61.24{\scriptsize$\pm$1.12}/44.32{\scriptsize$\pm$2.43} & 57.61{\scriptsize$\pm$1.27}/41.02{\scriptsize$\pm$1.18} \\
        & & 30 & 79.94{\scriptsize$\pm$0.63}/74.89{\scriptsize$\pm$1.27} & 76.78{\scriptsize$\pm$1.15}/69.54{\scriptsize$\pm$1.94} & 71.72{\scriptsize$\pm$1.98}/61.35{\scriptsize$\pm$2.79} & 68.03{\scriptsize$\pm$1.71}/54.57{\scriptsize$\pm$2.97} \\ \hline
        \multirow{4}{*}{CReST} & \multirow{2}{*}{+ Early fusion} & 10 & 85.45{\scriptsize$\pm$0.08}/81.09{\scriptsize$\pm$0.19} & 82.76{\scriptsize$\pm$0.20}/76.38{\scriptsize$\pm$0.30} & 75.71{\scriptsize$\pm$0.21}/61.83{\scriptsize$\pm$0.37} & 67.07{\scriptsize$\pm$0.17}/48.28{\scriptsize$\pm$0.51} \\
        & & 30 & 90.49{\scriptsize$\pm$0.11}/86.15{\scriptsize$\pm$0.23} & 87.74{\scriptsize$\pm$0.20}/81.27{\scriptsize$\pm$0.38} & 83.01{\scriptsize$\pm$0.19}/72.54{\scriptsize$\pm$0.43} & 75.50{\scriptsize$\pm$0.28}/58.62{\scriptsize$\pm$0.57} \\ \cline{2-7}
        & \multirow{2}{*}{+ Late fusion} & 10 & 72.10{\scriptsize$\pm$0.09}/65.63{\scriptsize$\pm$0.18} & 69.77{\scriptsize$\pm$0.12}/60.80{\scriptsize$\pm$0.30} & 64.28{\scriptsize$\pm$0.21}/50.29{\scriptsize$\pm$0.30} & 60.77{\scriptsize$\pm$0.15}/43.92{\scriptsize$\pm$0.29} \\
        & & 30 & 80.59{\scriptsize$\pm$0.11}/74.97{\scriptsize$\pm$0.20} & 78.55{\scriptsize$\pm$0.17}/70.18{\scriptsize$\pm$0.39} & 73.65{\scriptsize$\pm$0.17}/62.33{\scriptsize$\pm$0.30} & 68.43{\scriptsize$\pm$0.20}/52.55{\scriptsize$\pm$0.42} \\ \hline
        \multirow{4}{*}{Adsh} & \multirow{2}{*}{+ Early fusion} & 10 & 78.77{\scriptsize$\pm$0.50}/78.20{\scriptsize$\pm$0.33} & 74.95{\scriptsize$\pm$0.42}/71.44{\scriptsize$\pm$0.41} & 65.48{\scriptsize$\pm$0.82}/57.29{\scriptsize$\pm$0.85} & 57.65{\scriptsize$\pm$0.60}/42.18{\scriptsize$\pm$0.58} \\
        & & 30 & 90.06{\scriptsize$\pm$0.06}/85.97{\scriptsize$\pm$0.12} & 88.25{\scriptsize$\pm$1.48}/81.77{\scriptsize$\pm$0.45} & 82.36{\scriptsize$\pm$0.29}/72.04{\scriptsize$\pm$0.75} & 72.92{\scriptsize$\pm$0.29}/58.66{\scriptsize$\pm$0.85} \\ \cline{2-7}
        & \multirow{2}{*}{+ Late fusion} & 10 & 62.35{\scriptsize$\pm$0.75}/63.12{\scriptsize$\pm$1.42} & 58.36{\scriptsize$\pm$1.13}/56.74{\scriptsize$\pm$1.14} & 55.49{\scriptsize$\pm$0.46}/47.31{\scriptsize$\pm$0.73} & 51.43{\scriptsize$\pm$0.37}/41.58{\scriptsize$\pm$0.45} \\
        & & 30 & 80.27{\scriptsize$\pm$0.49}/74.86{\scriptsize$\pm$0.26} & 77.32{\scriptsize$\pm$0.66}/69.71{\scriptsize$\pm$1.29} & 72.13{\scriptsize$\pm$0.12}/58.38{\scriptsize$\pm$0.20} & 68.77{\scriptsize$\pm$0.14}/51.70{\scriptsize$\pm$0.34} \\ \hline
        \multicolumn{2}{c}{\multirow{2}{*}{SSMVAE-CI}} & 10 & 91.69{\scriptsize$\pm$0.25}/86.98{\scriptsize$\pm$0.41} & 87.85{\scriptsize$\pm$0.16}/80.33{\scriptsize$\pm$0.31} & 82.93{\scriptsize$\pm$0.80}/71.32{\scriptsize$\pm$1.53} & 74.79{\scriptsize$\pm$0.54}/59.07{\scriptsize$\pm$0.50} \\
        \multicolumn{2}{c}{} & 30 & 92.68{\scriptsize$\pm$0.53}/88.79{\scriptsize$\pm$0.89} & 89.72{\scriptsize$\pm$0.51}/83.94{\scriptsize$\pm$0.76} & 85.61{\scriptsize$\pm$0.34}/76.12{\scriptsize$\pm$0.87} & 79.47{\scriptsize$\pm$0.61}/64.44{\scriptsize$\pm$1.53} \\ \hline
    \end{tabular}}
\end{sc}
\end{small}
\end{center}
\vskip -0.1in
\end{table}

The table demonstrates the superior performance of the proposed model across both overall accuracy and minority-class accuracy, with performance gains becoming more pronounced at higher imbalance ratios.  
In particular, SSMVAE-CI consistently outperformed CReST and Adsh---regardless of whether early or late fusion was applied---highlighting the benefit of natively modeling multiple data modalities rather than relying on external multimodal fusion strategies.
While CMML and SMLN showed competitive results at lower imbalance levels, their performance, particularly on minority classes, degraded significantly as the imbalance ratio increases due to their lack of class-imbalance handling. Interestingly, even the baseline SSVAE, which was designed only for semi-supervised learning, achieved the second-best performance in terms of both overall and minority-class accuracy when $\beta = 10\%$. This suggests that generative models may generalize better than discriminative models in imbalanced training scenarios, especially under limited supervision.

The performance advantage of the proposed model is further attributed to its effective use of unlabeled data to mitigate class imbalance, as reflected in the performance gap between SSMVAE-CI and the baseline methods. 
Specifically, since GAMO and LDAM cannot utilize unlabeled data, their performance suffered as the imbalance ratio $r$ increases and the labeled data ratio $\beta$ decreased.  For example, when $\beta = 30\%$, SSMVAE-CI exceeded LDAM with early fusion by $3.58\%$ in overall accuracy and $5.98\%$ in minority-class accuracy. These margins increased to $8.24\%$ and $12.28\%$, respectively, when $\beta = 10\%$. 
Moreover, the gap between SSMVAE-CI and CReST or Adsh (with both fusion strategies) increased as $r$ increased, validating the robustness of the proposed model under class-imbalanced settings. For instance, when $\beta=10\%$, the gains over CReST with early fusion were $6.24\%$ (overall) and $5.89\%$ (minority) for $r=10$, which increased to $7.72\%$ and $10.79\%$, respectively, for $r=100$, further demonstrating the effectiveness of SSMVAE-CI in addressing severe class imbalance.

\subsection{Results on UPMC Food-101 Dataset}\label{sec:food101}

\begin{table}[t]
\caption{Overall accuracies/minority-class accuracies for the UPMC Food-101 dataset.}
\label{table:accuracy_food101}
\vskip 0.15in
\begin{center}
\begin{small}
\begin{sc}
    \resizebox{\textwidth}{!}{\begin{tabular}{cccccc}
        \hline
        \multicolumn{2}{c}{Model} & $r=10$ & $r=20$ & $r=50$ & $r=100$ \\ \hline
        \multicolumn{2}{c}{CMML} & 17.06{\scriptsize$\pm$0.34}/13.19{\scriptsize$\pm$0.37} & 14.89{\scriptsize$\pm$0.37}/6.24{\scriptsize$\pm$0.28} & 9.92{\scriptsize$\pm$0.15}/1.97{\scriptsize$\pm$0.23} & 8.66{\scriptsize$\pm$0.90}/0.96{\scriptsize$\pm$0.29} \\ \hline
        \multicolumn{2}{c}{SMLN} & 3.21{\scriptsize$\pm$0.07}/0.82{\scriptsize$\pm$0.21} & 3.21{\scriptsize$\pm$0.21}/0.80{\scriptsize$\pm$0.18} & 3.17{\scriptsize$\pm$0.11}/0.79{\scriptsize$\pm$0.17} & 3.08{\scriptsize$\pm$0.09}/0.77{\scriptsize$\pm$0.13} \\ \hline
        \multirow{2}{*}{SSVAE} & + Early fusion & 29.64{\scriptsize$\pm$1.71}/18.63{\scriptsize$\pm$1.93} & 22.57{\scriptsize$\pm$1.52}/11.21{\scriptsize$\pm$2.12} & 17.31{\scriptsize$\pm$0.78}/5.26{\scriptsize$\pm$1.17} & 14.12{\scriptsize$\pm$1.03}/3.01{\scriptsize$\pm$1.16} \\ \cline{2-6}
        & + Late fusion & 28.58{\scriptsize$\pm$1.24}/17.92{\scriptsize$\pm$1.57} & 20.77{\scriptsize$\pm$1.33}/10.42{\scriptsize$\pm$1.95} & 15.86{\scriptsize$\pm$0.96}/4.99{\scriptsize$\pm$1.03} & 13.22{\scriptsize$\pm$0.91}/2.88{\scriptsize$\pm$1.05} \\ \hline
        \multirow{2}{*}{GAMO} & + Early fusion & 2.07{\scriptsize$\pm$0.04}/0.45{\scriptsize$\pm$0.05} & 2.05{\scriptsize$\pm$0.03}/0.50{\scriptsize$\pm$0.08} & 1.97{\scriptsize$\pm$0.05}/0.48{\scriptsize$\pm$0.07} & 2.01{\scriptsize$\pm$0.04}/0.53{\scriptsize$\pm$0.05} \\ \cline{2-6}
        & + Late fusion & 2.06{\scriptsize$\pm$0.05}/0.50{\scriptsize$\pm$0.05} & 2.05{\scriptsize$\pm$0.06}/0.48{\scriptsize$\pm$0.07} & 2.01{\scriptsize$\pm$0.07}/0.49{\scriptsize$\pm$0.09} & 1.99{\scriptsize$\pm$0.04}/0.52{\scriptsize$\pm$0.07} \\ \hline
        \multirow{2}{*}{LDAM} & + Early fusion & 33.47{\scriptsize$\pm$1.68}/24.29{\scriptsize$\pm$1.35} & 25.24{\scriptsize$\pm$0.45}/14.22{\scriptsize$\pm$0.75} & 18.76{\scriptsize$\pm$0.35}/7.60{\scriptsize$\pm$0.86} & 16.19{\scriptsize$\pm$0.53}/4.11{\scriptsize$\pm$0.75} \\ \cline{2-6}
        & + Late fusion & 31.53{\scriptsize$\pm$0.25}/23.44{\scriptsize$\pm$0.36} & 23.77{\scriptsize$\pm$0.51}/13.84{\scriptsize$\pm$0.91} & 16.09{\scriptsize$\pm$0.60}/6.79{\scriptsize$\pm$1.17} & 15.03{\scriptsize$\pm$0.87}/3.37{\scriptsize$\pm$1.65} \\ \hline
        \multirow{2}{*}{CReST} & + Early fusion & 33.84{\scriptsize$\pm$0.80}/22.01{\scriptsize$\pm$0.94} & 29.07{\scriptsize$\pm$0.73}/14.81{\scriptsize$\pm$0.73} & 19.18{\scriptsize$\pm$1.84}/6.57{\scriptsize$\pm$1.86} & 17.42{\scriptsize$\pm$0.38}/3.04{\scriptsize$\pm$0.28} \\ \cline{2-6}
        & + Late fusion & 32.09{\scriptsize$\pm$0.27}/20.68{\scriptsize$\pm$0.48} & 27.80{\scriptsize$\pm$0.33}/12.82{\scriptsize$\pm$0.42} & 17.17{\scriptsize$\pm$0.42}/5.20{\scriptsize$\pm$0.70} & 16.62{\scriptsize$\pm$0.42}/3.47{\scriptsize$\pm$0.62} \\ \hline
        \multirow{2}{*}{Adsh} & + Early fusion & 29.09{\scriptsize$\pm$0.21}/20.58{\scriptsize$\pm$0.66} & 21.32{\scriptsize$\pm$0.40}/10.21{\scriptsize$\pm$0.36} & 17.49{\scriptsize$\pm$0.29}/6.26{\scriptsize$\pm$0.21} & 14.72{\scriptsize$\pm$0.33}/3.64{\scriptsize$\pm$0.41} \\ \cline{2-6}
        & + Late fusion & 28.66{\scriptsize$\pm$0.32}/19.32{\scriptsize$\pm$0.50} & 20.91{\scriptsize$\pm$0.26}/9.75{\scriptsize$\pm$0.33} & 16.39{\scriptsize$\pm$0.40}/5.45{\scriptsize$\pm$0.55} & 13.47{\scriptsize$\pm$0.28}/2.91{\scriptsize$\pm$0.33} \\ \hline
        \multicolumn{2}{c}{SSMVAE-CI} & 49.97{\scriptsize$\pm$0.81}/38.36{\scriptsize$\pm$0.88} & 34.81{\scriptsize$\pm$1.01}/18.64{\scriptsize$\pm$1.13} & 28.93{\scriptsize$\pm$0.93}/10.65{\scriptsize$\pm$1.20} & 20.22{\scriptsize$\pm$0.55}/5.97{\scriptsize$\pm$0.65} \\ \hline
    \end{tabular}}
\end{sc}
\end{small}
\end{center}
\vskip -0.1in
\end{table}

To perform classification using the UPMC Food-101 dataset, we first preprocessed the textual descriptions and vectorized them using bag-of-words representation. During this process, we ensured that all words appearing in the dataset were included in the vector, resulting in a 15,064-dimensional bag-of-words vector for each sample. We also rescaled the images, which hand inconsistent sizes, so that all images had a side length of 128 pixels.

As before, we repeated the experiment 10 times, but with a fixed value of $\beta=30\%$ and randomly sampled, without replacement, the required number of labeled and unlabeled samples for each class in each experiment. The performance of the proposed model and the competing models is presented in Table \ref{table:accuracy_food101}, which reaffirms that our model outperforms the baseline models, especially when the dataset is severely imbalanced. Specifically, LDAM, CReST, and Adsh, with either early or late fusion, produced lower overall and minority-class accuracy than SSMVAE-CI across all imbalance ratios, reflecting their limited capacity to effectively handle multimodal data. CMML showed poor performance, with both  overall and minority-class accuracy approximately half that of other methods, likely due to its lack of mechanisms to address class imbalance. Interestingly, SSVAE with early and late fusion, despite sharing the same objective as CMML, achieved competitive results relative to other baselines, further demonstrating the flexibility and robustness of generative models. We excluded the comparisons with SMLN and GAMO combined with early and late fusion, as they failed to learn on the UPMC Food-101 dataset.

\subsection{Results on CMU-MOSEI Dataset}\label{sec:mosei}
To evaluate the performance of the proposed model on a more complex multimodal dataset involving more than two modalities, we conducted a series of experiments on the CMU-MOSEI dataset. This dataset includes three modalities: video, audio, and text. Each video is segmented into multiple short utterances, each annotated with a sentiment score ranging from -3 to 3, where -3 indicates strongly negative sentiment and 3 indicates strongly positive sentiment. Our task was to predict the sentiment scores based on the three modalities. Following the procedure described in \cite{liang2021multibench}, we split the dataset into training and test sets and extracted features from the three modalities. These features, represented as tensors, were used as inputs to both the proposed model and the competing baselines. We did not perform any additional adjustments to class sizes, as the dataset is naturally imbalanced.

The experimental results are presented in Table \ref{table:accuracy_mosei}. Due to the dataset’s complexity, size, and noisy features, the performance differences between the proposed model and competing methods are less pronounced. Nevertheless, the proposed model generally outperforms the baselines, with the largest observed gaps in overall and minority-class accuracy reaching $14.48\%$ and $14.70\%$, respectively, for $\beta=10\%$, and $13.35\%$ and $15.72\%$, respectively, for $\beta=30\%$. Notably, while CMML and SMLN achieved higher overall accuracy than GAMO, they exhibited lower minority-class accuracy, indicating that they perform well on majority classes but struggle with minority-class predictions due to their lack of mechanisms for handling class imbalance. Consistent with results on other datasets, SSVAE combined with early or late fusion demonstrated strong overall performance and ranked second in minority-class accuracy, further highlighting the advantage of generative models over discriminative models.

\begin{table}[htbp]
\caption{Overall accuracies/minority-class accuracies for the CMU-MOSEI dataset.}
\label{table:accuracy_mosei}
\vskip 0.15in
\begin{center}
\begin{small}
\begin{sc}
    \begin{tabular}{cccc}
        \hline
        \multicolumn{2}{c}{Model} & $\beta=10\%$ & $\beta=30\%$ \\ \hline
        \multicolumn{2}{c}{CMML} & 27.70{\scriptsize$\pm$2.92}/11.69{\scriptsize$\pm$2.78} & 28.46{\scriptsize$\pm$2.78}/12.70{\scriptsize$\pm$2.57} \\ \hline
        \multicolumn{2}{c}{SMLN} & 26.57{\scriptsize$\pm$0.17}/9.98{\scriptsize$\pm$0.52} & 27.95{\scriptsize$\pm$0.31}/10.42{\scriptsize$\pm$0.78} \\ \hline
        \multirow{2}{*}{SSVAE} & + Early fusion & 30.03{\scriptsize$\pm$3.08}/20.43{\scriptsize$\pm$7.73} & 33.10{\scriptsize$\pm$2.27}/21.21{\scriptsize$\pm$6.03} \\ \cline{2-4}
        & + Late fusion & 28.37{\scriptsize$\pm$0.99}/20.31{\scriptsize$\pm$2.98} & 30.04{\scriptsize$\pm$1.77}/21.55{\scriptsize$\pm$4.08} \\ \hline
        \multirow{2}{*}{GAMO} & + Early fusion & 24.97{\scriptsize$\pm$1.47}/13.75{\scriptsize$\pm$2.91} & 25.28{\scriptsize$\pm$1.09}/14.58{\scriptsize$\pm$1.54} \\ \cline{2-4}
        & + Late fusion & 25.64{\scriptsize$\pm$0.67}/14.48{\scriptsize$\pm$1.47} & 26.88{\scriptsize$\pm$1.13}/13.09{\scriptsize$\pm$1.51} \\ \hline
        \multirow{2}{*}{LDAM} & + Early fusion & 33.60{\scriptsize$\pm$0.50}/13.74{\scriptsize$\pm$1.12} & 32.94{\scriptsize$\pm$0.58}/17.09{\scriptsize$\pm$1.93} \\ \cline{2-4}
        & + Late fusion & 33.09{\scriptsize$\pm$0.75}/13.56{\scriptsize$\pm$2.60} & 31.69{\scriptsize$\pm$0.71}/16.19{\scriptsize$\pm$1.02} \\ \hline
        \multirow{2}{*}{CReST} & + Early fusion & 30.65{\scriptsize$\pm$0.24}/16.25{\scriptsize$\pm$0.13} & 32.90{\scriptsize$\pm$0.34}/15.59{\scriptsize$\pm$0.45} \\ \cline{2-4}
        & + Late fusion & 29.16{\scriptsize$\pm$0.33}/15.86{\scriptsize$\pm$0.50} & 29.58{\scriptsize$\pm$0.66}/17.22{\scriptsize$\pm$1.43} \\ \hline
        \multirow{2}{*}{Adsh} & + Early fusion & 21.89{\scriptsize$\pm$0.15}/11.48{\scriptsize$\pm$0.32} & 25.22{\scriptsize$\pm$0.99}/13.80{\scriptsize$\pm$0.10} \\ \cline{2-4}
        & + Late fusion & 21.65{\scriptsize$\pm$0.18}/11.99{\scriptsize$\pm$1.34} & 25.58{\scriptsize$\pm$0.44}/14.03{\scriptsize$\pm$1.20} \\ \hline
        \multicolumn{2}{c}{SSMVAE-CI} & 36.13{\scriptsize$\pm$1.53}/24.68{\scriptsize$\pm$1.71} & 38.57{\scriptsize$\pm$1.32}/26.14{\scriptsize$\pm$1.85} \\ \hline
    \end{tabular}
\end{sc}
\end{small}
\end{center}
\vskip -0.1in
\end{table}

\subsection{Evaluation Under Missing Modalities}
In real-world multimodal datasets, it is common for some samples—either labeled or unlabeled—to lack one or more modalities. This may occur due to data corruption, limited sensor availability, or collection constraints in practical settings. A straightforward approach to handle such missing data is to discard incomplete samples and train models only on samples with all modalities. However, this can significantly reduce the available training data and result in degraded performance due to the loss of valuable information.

To address this issue, the proposed model utilizes samples with missing modalities as additional training signals, rather than discarding them. This capability is enabled by incorporating a Student’s $t$-distribution prior. Due to its heavy-tailed nature, the $t$-prior encourages latent representations to occupy broader regions of the latent space with high probability density, allowing the model to learn more diverse and expressive features. This, in turn, facilitates the implicit estimation of missing modalities during training, ultimately improving predictive performance.

To evaluate the effectiveness of SSMVAE-CI under missing modality scenarios, we conducted experiments on the MNIST-SVHN dataset. 
We artificially generated a modified version of the dataset by randomly removing one modality for all samples in the first half of the training set. When training SSMVAE-CI, its modality-specific encoders allowed computation of the loss only over observed modalities, thus enabling the model to fully utilize all samples. In contrast, since baseline models require complete modality information, we trained them solely on the second half of the dataset that retained both modalities.
The results of the experiments with $r=50$ and $\beta=10\%$ are presented in Table \ref{table:missing}. For SSMVAE-CI, we also report results from an additional experiment in which the model was trained on the second half of the dataset containing both modalities, to enable a fair comparison with the baseline models. In the table, the decrease in accuracy relative to the original dataset (Table \ref{table:accuracy_benchmarks}) is shown in  parentheses.

\begin{table}[htbp]
\caption{Overall accuracies/minority-class accuracies for the MNIST-SVHN dataset with missing modalities.}
\label{table:missing}
\vskip 0.15in
\begin{center}
\begin{small}
\begin{sc}
    \begin{tabular}{cccccc}
        \hline
        \multicolumn{2}{c}{Model} & \multicolumn{2}{c}{Overall} & \multicolumn{2}{c}{Minority} \\ \hline
        \multicolumn{2}{c}{CMML} & 68.69{\scriptsize$\pm$0.87} & ($-4.93$) & 41.76{\scriptsize$\pm$0.59} & ($-9.86$) \\ \hline
        \multicolumn{2}{c}{SMLN} & 59.34{\scriptsize$\pm$1.19} & ($-8.79$) & 35.05{\scriptsize$\pm$1.57} & ($-11.29$) \\ \hline
        \multirow{2}{*}{SSVAE} & + Early fusion & 71.31{\scriptsize$\pm$0.23} & ($-7.51$) & 51.63{\scriptsize$\pm$0.73} & ($-12.96$) \\ \cline{2-6}
        & + Late fusion & 61.60{\scriptsize$\pm$0.35} & ($-4.44$) & 44.71{\scriptsize$\pm$0.70} & ($-5.58$) \\ \hline
        \multirow{2}{*}{GAMO} & + Early fusion & 64.32{\scriptsize$\pm$0.59} & ($-7.32$) & 38.80{\scriptsize$\pm$0.80} & ($-11.65$) \\ \cline{2-6}
        & + Late fusion & 52.72{\scriptsize$\pm$0.45} & ($-4.59$) & 32.17{\scriptsize$\pm$0.58} & ($-4.65$) \\ \hline
        \multirow{2}{*}{LDAM} & + Early fusion & 65.73{\scriptsize$\pm$0.59} & ($-7.78$) & 41.11{\scriptsize$\pm$1.04} & ($-13.61$) \\ \cline{2-6}
        & + Late fusion & 53.44{\scriptsize$\pm$0.57} & ($-7.80$) & 29.58{\scriptsize$\pm$1.21} & ($-14.74$) \\ \hline
        \multirow{2}{*}{CReST} & + Early fusion & 68.05{\scriptsize$\pm$0.84} & ($-7.66$) & 46.27{\scriptsize$\pm$1.73} & ($-15.56$) \\ \cline{2-6}
        & + Late fusion & 58.01{\scriptsize$\pm$0.18} & ($-6.27$) & 42.46{\scriptsize$\pm$0.84} & ($-7.83$) \\ \hline
        \multirow{2}{*}{Adsh} & + Early fusion & 61.42{\scriptsize$\pm$0.63} & ($-4.06$) & 43.17{\scriptsize$\pm$1.02} & ($-14.12$) \\ \cline{2-6}
        & + Late fusion & 50.48{\scriptsize$\pm$0.61} & ($-5.01$) & 40.48{\scriptsize$\pm$0.89} & ($-6.83$) \\ \hline
        \multicolumn{2}{c}{SSMVAE-CI (using half)} & 78.30{\scriptsize$\pm$0.29} & ($-4.63$) & 65.42{\scriptsize$\pm$0.94} & ($-5.90$) \\ \hline
        \multicolumn{2}{c}{SSMVAE-CI (using all)} & 80.85{\scriptsize$\pm$0.65} & ($-2.08$) & 68.88{\scriptsize$\pm$1.10} & ($-2.44$) \\ \hline
    \end{tabular}
\end{sc}
\end{small}
\end{center}
\vskip -0.1in
\end{table}

As shown in the table, the proposed model achieved significantly higher accuracies than the baseline models, with gains of at least $6.99\%$ in overall accuracy and $13.79\%$ in minority-class accuracy. More importantly, the performance drop—reported in parentheses—from the original (complete-modality) dataset provides key insights. While all models experienced accuracy degradation due to the missing modality in half of the dataset, SSMVAE-CI trained on the full dataset (including both complete and incomplete samples) exhibited the smallest drop: just $2.08\%$ in overall accuracy and $2.44\%$ in minority-class accuracy. Notably, this drop is smaller than that of SSMVAE-CI trained only on the complete half of the dataset, highlighting the model’s ability to exploit the additional information from samples with missing modalities—an advantage not shared by baseline models, which cannot handle such incomplete data.

\subsection{Ablation Study}
We conducted an ablation study on the MNIST-SVHN dataset to investigate the individual contributions of key components in the proposed method. While this dataset was chosen for its simplicity, similar trends were observed across the other two datasets. 
The proposed method, SSMVAE-CI, extends the vanilla SSVAE by incorporating two major enhancements: a PoE framework and the use of Student’s $t$-distributions for the prior, encoder, and decoder. To assess the impact of these components, we began with the vanilla SSVAE using either early or late fusion, and progressively added each component, tracking changes in both overall and minority-class accuracy. The results for two extreme imbalance ratios, $r = 50$ and $r = 100$, with a fixed $\beta = 10\%$, are presented in Table \ref{table:ablation}.

The key findings from the results in Table \ref{table:ablation} are summarized as follows: (1) Replacing the Student’s $t$-distributions in the prior, encoder, and decoder with Gaussian distributions led to a noticeable drop in classification accuracy, as Gaussian distributions lack the heavy-tailed property that enables better handling of class-imbalanced data. (2) Substituting the PoE framework with simple early or late fusion resulted in reduced performance, indicating that such na\"ive fusion strategies are insufficient for capturing inter-modal correlations effectively. (3) Performance declined even further when using the vanilla SSVAE with either fusion strategy, as it lacks both the PoE mechanism and the heavy-tailed modeling capability of $t$-distributions, making it less suitable for class-imbalanced multimodal learning.

\begin{table}[ht]
\caption{Ablation study for SSMVAE-CI on the MNIST-SVHN dataset.}
\label{table:ablation}
\vskip 0.15in
\begin{center}
\begin{small}
\begin{sc}
    \resizebox{\textwidth}{!}{\begin{tabular}{lcc}
    \hline
    \multirow{2}{*}{Ablation study} & $r=50$, $\beta=10\%$ & $r=100$, $\beta=10\%$ \\
    & Overall/Minority & Overall/Minority \\ \hline
    \begin{tabular}[c]{@{}l@{}}SSMVAE-CI\\ \quad $\Rightarrow$ SSVAE + PoE + Student's $t$-distributions\end{tabular} & 82.93/71.32 & 74.79/59.07 \\
    \begin{tabular}[c]{@{}l@{}}Replacing Student's $t$-distributions with Gaussian distributions\\ \quad $\Rightarrow$ SSVAE + PoE + Gaussian distributions\end{tabular} & 80.83/65.72 & 71.55/53.32 \\
    \begin{tabular}[c]{@{}l@{}}Replacing the PoE structure with an early fusion\\ \quad $\Rightarrow$ SSVAE + Early fusion + Student's $t$-distributions\end{tabular} & 81.72/67.30 & 72.72/55.51 \\
    \begin{tabular}[c]{@{}l@{}}Replacing the PoE structure with a late fusion\\ \quad $\Rightarrow$ SSVAE + Late fusion + Student's $t$-distributions\end{tabular} & 72.39/56.71 & 66.20/48.82 \\
    \begin{tabular}[c]{@{}l@{}}SSVAE combined with an early fusion\\ \quad $\Rightarrow$ SSVAE + Early fusion + Gaussian distributions\end{tabular} & 76.92/62.81 & 69.36/51.96 \\
    \begin{tabular}[c]{@{}l@{}}SSVAE combined with a late fusion\\ \quad $\Rightarrow$ SSVAE + Late fusion + Gaussian distributions\end{tabular} & 66.04/50.29 & 60.50/40.49 \\ \hline
    \end{tabular}}
\end{sc}
\end{small}
\end{center}
\vskip -0.1in
\end{table}

\section{Conclusion}\label{sec:conclusion}
In this paper, we proposed SSMVAE-CI, a novel model for semi-supervised learning with class-imbalanced multimodal datasets. We designed a generative model with an inference structure  consisting of modality-specific encoders that share a  latent variable across all modalities. This approach approximates the joint posterior as a product of individual posteriors with a prior expert, effectively reducing computational complexity. Additionally, we replaced the typical Gaussian distribution used in the prior, encoder, and decoder with the Student's $t$-distribution to better account for the heavy-tailed nature of class-imbalanced datasets.

To validate our model, we conducted experiments using a combination of two benchmark image datasets and two real-world datasets containing diverse modalities. Our results demonstrate that the proposed model is well-suited for classifying samples from class-imbalanced multimodal datasets with partial supervision. Beyond the evaluated datasets, the proposed framework has strong potential in domains such as healthcare and finance, where multimodal, partially labeled, and highly imbalanced data are common. For example, in medical diagnosis, rare conditions appear infrequently and span heterogeneous modalities such as medical imaging, clinical notes, and physiological signals, making balanced annotation difficult. Similarly, in financial risk assessment, fraudulent behaviors constitute minority patterns embedded in multimodal sources such as transaction histories and textual records. By jointly addressing class imbalance, multimodality, and partial supervision within a unified generative framework, SSMVAE-CI can facilitate more reliable, data-efficient, and equitable decision-making in such high-stakes real-world scenarios.

In future work, we plan to extend our model to accommodate distribution mismatches between the labeled and unlabeled sets, where the distribution of the unlabeled set is often unknown in practice. This extension would expand the range of datasets the model can handle, thus enhancing its applicability to a wider array of real-world problems.

\section*{Supplementary Materials}
\textbf{Supplementary document}: The pdf file contains the full derivation of the objective function used for SSMVAE-CI.

\noindent\textbf{Python code}: ``Codes.zip" contains the codes and datasets used in this paper.

\section*{Acknowledgments}
The authors would like to thank the referees, the associate editor, and the editor for reviewing this article and providing valuable comments. 
This work was supported by Samsung Electronics Co., Ltd. and the National Research Foundation of Korea (NRF) grant funded by the Korea government (MSIT) (2023R1A2C2005453, RS-2023-00218913).

\section*{Conflict of Interest Disclosure}
The authors declare that they have no conflict of interests that could have appeared to influence the work reported in this manuscript.

%\clearpage

\bibliographystyle{asa}
\bibliography{SSMVAE_reference}

\end{document}